%% file: colm2025_conference.tex
\documentclass{article} 
\usepackage[preprint]{colm2025_conference}

\usepackage{microtype}
\usepackage{hyperref}
\usepackage{url}
\usepackage{booktabs}

\usepackage{amsmath,amssymb}
\usepackage{graphicx}
\usepackage{tabularray}    
\usepackage{xcolor}        

\UseTblrLibrary{booktabs} 
\UseTblrLibrary{color}    

\usepackage{multirow} 
\usepackage{float} 
\usepackage[most]{tcolorbox} 
\usepackage[ruled,vlined]{algorithm2e}
\usepackage{lineno}

\definecolor{darkblue}{rgb}{0, 0, 0.5}
\hypersetup{colorlinks=true, citecolor=darkblue, linkcolor=darkblue, urlcolor=darkblue}

\title{\makebox[\textwidth][c]{Entropy-Based Adaptive Weighting for Self-Training}}

\author{
\makebox[\textwidth][c]{Xiaoxuan Wang \quad Yihe Deng \quad Mingyu Derek Ma \quad Wei Wang} \\
\makebox[\textwidth][c]{University of California Los Angeles}
}

\begin{document}

\ifcolmsubmission
\linenumbers
\fi

\maketitle

\begin{abstract}
The mathematical problem-solving capabilities of large language models have become a focal point of research, with growing interests in leveraging self-generated reasoning paths as a promising way to refine and enhance these models. These paths capture step-by-step logical processes while requiring only the correct answer for supervision.
The self-training method has been shown to be effective in reasoning tasks while eliminating the need for external models and manual annotations.
However, optimizing the use of self-generated data for model training remains an open challenge.
In this work, we propose \textbf{E}ntropy-Based \textbf{A}daptive Weighting for \textbf{S}elf-\textbf{T}raining (EAST), an adaptive weighting strategy designed to prioritize uncertain data during self-training. Specifically, EAST employs a mapping function with a tunable parameter that controls the sharpness of the weighting, assigning higher weights to data where the model exhibits greater uncertainty. This approach guides the model to focus on more informative and challenging examples, thereby enhancing its reasoning ability.
We evaluate our approach on GSM8K and MATH benchmarks. Empirical results show that, while the vanilla method yields virtually no improvement (0\%) on MATH, EAST achieves around a 1\% gain over backbone model. On GSM8K, EAST attains a further 1–2\% performance boost compared to the vanilla method. Our codebase is publicly available on GitHub\footnote{GitHub Link: https://github.com/mandyyyyii/east. Correspondence: xw27@g.ucla.edu}.
\end{abstract}

\input{section/introduction}

\input{section/preliminaries}

\input{section/method}

\input{section/experiment}
\input{section/related_work}
\input{section/conclusion}

\bibliography{colm2025_conference}
\bibliographystyle{colm2025_conference}

\include{section/appendix}

\end{document}

%% file: section/introduction.tex
\section{Introduction}
Mathematical reasoning is a key component of Large Language Model (LLM) capabilities, as it directly relates to logical consistency and problem-solving skills~\citep{yu2023metamath,zhang2024mathverse, gao2024omni, liu2024mathbench}. This area has drawn increasing attention because the correctness of a final mathematical answer can provide a direct, verifiable reward signal for reinforcement learning (RL) approaches, enabling LLM-generated reasoning paths for both self-training~\citep{zelikman2022star, singh2023beyond, xiong2024iterative} and distillation~\citep{ho2022large, fu2023specializing, gou2023tora}.

The core idea of both self-training and distillation is based on rejection sampling: for each given question, the LLM generates multiple responses and selects the reasoning paths that yield correct answers as positive samples for subsequent fine-tuning~\citep{zelikman2022star,singh2023beyond, luong2024reft}. Through iterative application of this self-training process, the LLM progressively enhances its performance. Recent studies have explored leveraging negative samples to construct preference pairs for reward models~\citep{hosseini2024v} or directly applying pair-wise alignment methods for fine-tuning~\citep{xu2024dpo, sun2024easy, zhong2024dpo, ivison2024unpacking, saeidi2024insights, xiong2024iterative}. 

However, many self-training methodologies treat generated data uniformly, assigning equal importance to all generated examples. Such approaches may overlook the varying educational value of different data points, which can potentially impede the model’s ability to prioritize the most informative data and possibly limits its overall learning effectiveness. This observation raises a question: could reweighting training data during self-training improve reasoning capabilities? If so, which data should be prioritized, and to what extent should it be emphasized?



In the self-training pipeline for reasoning tasks, additional training on already well-understood questions brings minimal gains and risks overfitting the model to simpler data. Instead, focusing on challenging questions—where the model struggles—promises more efficient learning~\citep{huang2022large,singh2023beyond}. Moreover, large language models can exhibit resistance to updating their predictions, particularly in cases where they demonstrate high confidence. In contrast, guiding the model to focus on areas of uncertainty enhances its training effectiveness~\citep{kumar2024training, li2024confidence}. 

To address this gap, we introduce \textbf{E}ntropy-Based \textbf{A}daptive Weighting for \textbf{S}elf-\textbf{T}raining (EAST), a novel method that assigns adaptive weights to training data during self-training based on model uncertainty, measured via the entropy of the model’s sample distribution for a given question. Specifically, given multiple samples generated by an LLM for a question, EAST clusters these samples by their final answers and computes the entropy over the resulting cluster-based distribution. EAST then applies a mapping function that transforms the entropy value into a bounded weight under predefined constraints. This function includes a tunable parameter that controls the sharpness of the weighting, allowing flexible emphasis on uncertain data. By assigning higher weights to high-entropy data—those reflecting greater model uncertainty—EAST encourages the model to focus on more informative and challenging examples during training. Prioritizing such examples not only enhances reasoning capability but also helps prevent overfitting to overconfident data. Moreover, EAST is a flexible framework that supports both iterative self-training and integration with various loss functions, making it broadly applicable across different training settings.
\begin{figure*}[t]
\vskip-1.5em
	\centering
	\includegraphics[width=\linewidth]{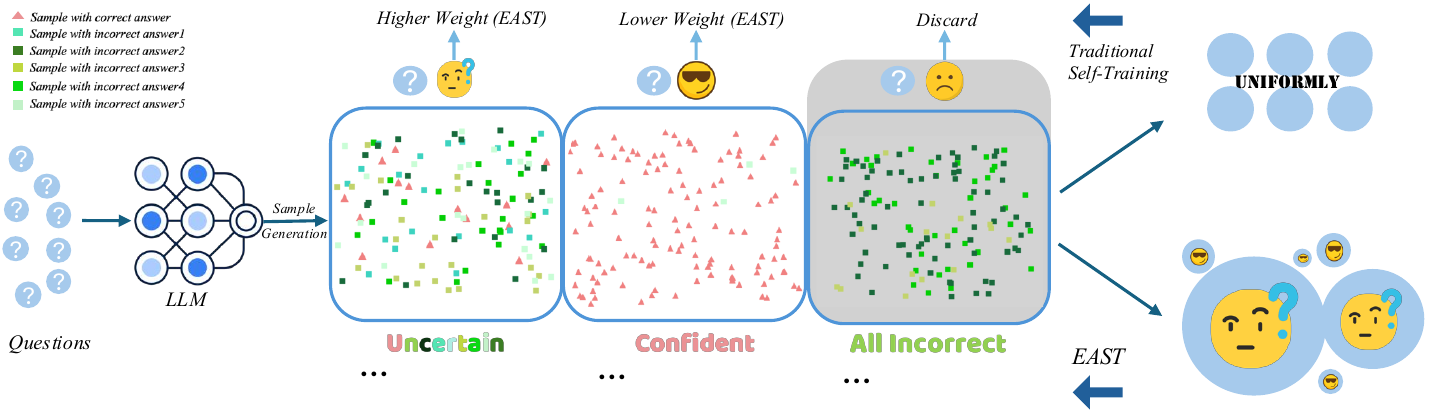}
     \vskip-0.5em 
	\caption{Comparison between the traditional self-training pipeline and EAST. The LLM generates $n$ responses per question, clustered by final answers. Questions with all incorrect answers are discarded. Self-training fine-tunes uniformly on the rest, while EAST assigns higher weights to questions with diverse (uncertain) answers and lower weights to consistent (confident) ones.}
	\label{fig:many}
 \vskip-1em
\end{figure*}


We evaluate EAST by incorporating it into SFT, DPO~\citep{rafailov2024direct}, and KTO~\citep{ethayarajh2024kto} loss functions on two mathematical reasoning benchmarks GSM8K~\citep{cobbe2021training} and MATH~\citep{hendrycks2021measuring}. EAST achieves notable performance gains of 5.6\% on GSM8K and approximately 1\% on MATH over the default backbone model, substantially outperforming vanilla SFT, which yields only a 3.9\% improvement on GSM8K and no gain on MATH. A similar trend is observed for both DPO and KTO, with performance improvements of up to 1.7\% on GSM8K and 2.1\% on MATH compared to the vanilla methods. We further show that EAST consistently surpasses vanilla method through iterative training. In addition, we demonstrate the effectiveness of entropy-based weighting, which outperforms other weighting strategies by better leveraging uncertain data and reducing reliance on overconfident data during training, thereby enhancing reasoning capabilities. 

Our contributions are summarized as follows: 
\begin{itemize}
\vskip-0.4em 
\item \textbf{Entropy-Based weighting}: a new weighting strategy that leverages uncertainty information, derived from the entropy of the model’s sample distribution over the training data
\item \textbf{Mapping function}: a novel mapping function that controls the extent to which higher uncertain data are weighted
\item \textbf{Experimental evaluation }: EAST further boosts self-training performance compared to the vanilla method. 
\end{itemize}

%% file: section/preliminaries.tex
\vspace{-0.5em}
\section{Preliminaries}
We consider a large language model (LLM) parameterized by \(\theta\), denoted \(p_\theta\). Given a prompt \(x = [x_1,\dots,x_n]\), the model generates a response \(y = [y_1,\dots,y_m]\) via an auto-regressive factorization:
\[
p_\theta(y \mid x) \;=\; \prod_{j=1}^{m} p_\theta\bigl(y_j \mid x,\, y_{<j}\bigr),
\]
where \(y_{<j} = [y_1,\dots,y_{j-1}]\).



\textbf{Self-Training Pipeline.}
Self-training addresses the scarcity of human-annotated data by leveraging the target model to generate completion paths~\citep{zelikman2022star, singh2023beyond,chen2024self}.
Formally, under mathematical context, given a dataset of input-output pairs $({(x_i, y_i)}_{i=1}^N)$, where $(x_i)$ represents a mathematical question and $(y_i)$ is its corresponding ground truth answer, we aim to introduce an intermediate reasoning path $(r_i)$ that delineates the logical steps from $(x_i)$ to $(y_i)$. Let $(p_\theta(r_i \mid x_i))$ denote the model's distribution over possible reasoning paths, parameterized by ($\theta$). The self-training process involves: 
\begin{enumerate}
\item Sampling reasoning paths $(\hat{r_i} \sim p_\theta(\cdot \mid x_i))$
\item Evaluating the correctness of $(\hat{r_i})$ by verifying if it leads to the ground truth ($y_i$)
\item Updating the training set with validated triples $({(x_i, y_i, \hat{r}_i)}_{i=1}^M)$
\item Updating model parameters ($\theta$) through iterative training
\end{enumerate}

For supervised fine-tuning (SFT), only sample paths that yield correct answers are incorporated into the training data. Alignment methods such as direct preference optimization (DPO, \citeauthor{rafailov2024direct}, \citeyear{rafailov2024direct}) utilize both correct and incorrect sample paths to learn from contrastive preferences, where correct paths serve as positive pairs and incorrect ones as negative pairs.


%% file: section/method.tex
\section{Method}
\begin{figure*}[t]
\vskip-0.3em
	\centering
	\includegraphics[width=\linewidth]{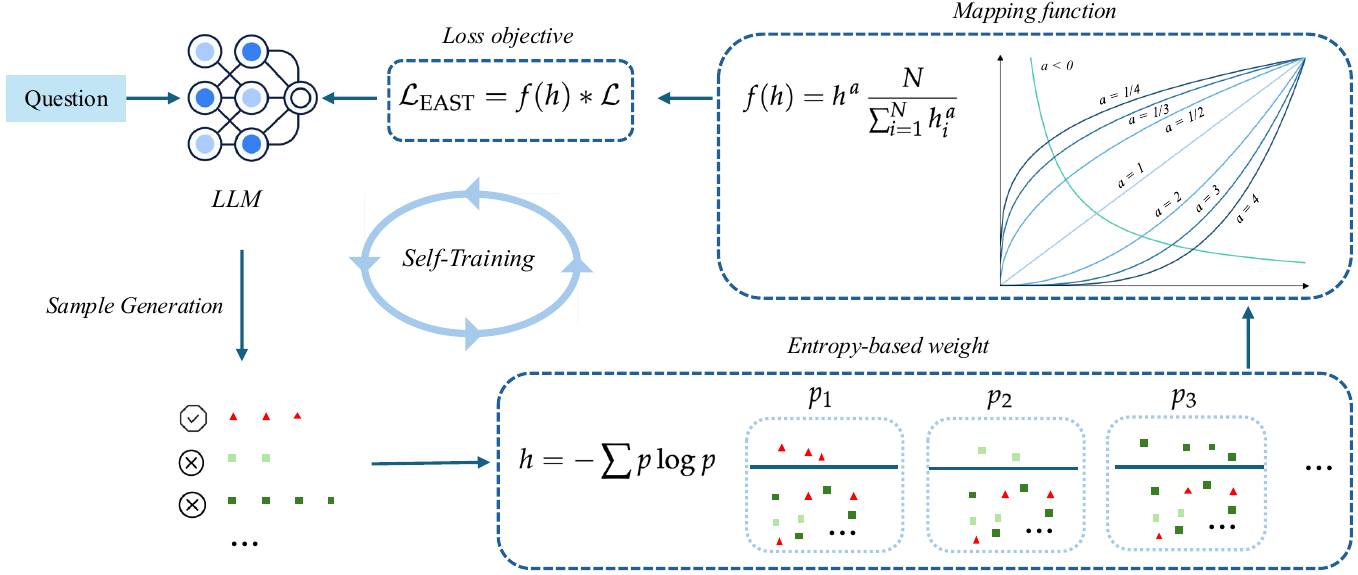}
    \vskip-0.07em
	\caption{The framework of EAST. For each training question, the LLM generates $n$ responses, clustered by final answers. Entropy value is computed from the cluster distribution, transformed via mapping function, and integrated as weight into the loss objective. }
	\label{fig:apo}
 \vskip-0.2em
\end{figure*}

In this section, we introduce EAST, a novel weighting method that prioritizes uncertain data within the self-training pipeline. We begin by presenting the entropy-based weighting strategy, followed by the proposed mapping function, and conclude with the final loss objective.
Figure~\ref{fig:many} demonstrates the comparison between the traditional self-training pipeline and EAST. Figure~\ref{fig:apo} represents the detailed framework of EAST. 

\subsection{Entropy-Based Weight}
Many studies have found that large language models (LLMs) tend to be resistant to changing their predictions, particularly when they are highly confident in their responses~\citep{kumar2024training, yang2024confidence, li2024confidence}. Therefore, guiding the model to focus on areas where it lacks confidence or is uncertain becomes a natural next step. Research indicates that prioritizing learning from uncertain questions—rather than those where the model is stubborn—leads to improved reasoning capabilities.~\citep{kumar2024training, li2024confidence}.

Based on this observation, we introduce an entropy-based weighting approach that encourages models to focus on learning from uncertain data. The key insight is that questions with higher entropy reflect greater uncertainty in the model’s predictions, indicating a lack of strong preference among possible answers. By prioritizing high-entropy questions during training, the model is encouraged to focus on informative and challenging examples, which enhances reasoning capabilities and helps prevent overfitting to overconfident examples. We further demonstrate the advantage of entropy-based weighting over alternative weighting strategies in Section~\ref{sec:ablation}, including accuracy-based weighting, which considers the proportion of correct answers (accuracy ratio), and rejection-based weighting, which captures the dominance of the most frequent incorrect answer (dominant incorrect ratio).

In the self-training pipeline, we generate $n$ samples for each question and cluster them based on their final answers, with each cluster representing a distinct answer. The number of clusters for a given question depends on the diversity of the model’s outputs, which we denote as $k_i$ (where $k_i \leq n$) for question $x_i$. The underlying assumption is that samples leading to the same final answer tend to share similar reasoning patterns. Thus, each cluster reflects the model's implicit preference for a particular reasoning path. A larger number of sparse clusters ($k_i$) indicates greater model uncertainty for that question, while a distribution concentrated in a single cluster suggests higher model confidence. The model's uncertainty for a given question $x_i$ is quantified through the entropy value, computed over the $k_i$ answer clusters as follows: 
\begin{equation}
\label{eq:entropy}
H(x_i) = -\sum_{j=1}^{k_i} p_j \log p_j
\end{equation}
where $p_j$ denotes the proportion of samples in cluster $j$ relative to the total number of samples for $x_i$. For simplicity, we denote $h_i=H(x_i)$. 

\subsection{Mapping Function}
\label{sec:mapping}
Given the entropy value $h_i$ on each question, our goal is to map this entropy value to a weight applied to the model loss using a function $f$. This mapping function $f$ must satisfy two constraints: (1) \textit{Non-negativity}—all transformed weights must be non-negative to ensure proper model training; (2) \textit{Normalization}—the transformed weights should have an average of 1 to prevent unintended effects on the learning rate, formally:
\begin{equation}
\min(f(h)) \geq 0, \quad \frac{1}{N} \sum_{i=1}^{N} f(h_i) = 1.
\end{equation}

\textbf{\textit{Attempt 1.}} A straightforward mapping function is the mean-division function:
\begin{equation}
    f(h) = \frac{h}{\mu}, \quad \mu = \frac{1}{N} \sum_{i=1}^{N} h_i
\end{equation}
which satisfies the normalization constraint. However, this approach lacks tunable parameters to control the distribution of transformed values.

\textbf{\textit{Attempt 2.}} To allow for control over the distribution of transformed values, we introduce a new parameter $R=f(\max(h)) - f(\min(h))$ that represents the range of mapped values. Therefore, instead of applying a fixed compression ratio(as in the mean-division function) to entropy values, we allow a controllable compression ratio $a$ that adapts the new output range $R$ and the original range $(\max(h) - \min(h))$:
\begin{equation}
    f(h) = a\,h + b, \quad a = \frac{R}{\max(h) - \min(h)}, \quad b = 1 - a \mu.
\end{equation}

Here, $b$ is determined by the normalization constraint: $\frac{1}{N} \sum_{i=1}^{N} (a h_i + b) = a \mu + b = 1$, which gives $b = 1 - a \mu$. The non-negativity constraint ($\min(f(h)) \geq 0$) requires $a (\min(h) - \mu) + 1 \geq 0$, yielding $R \leq \frac{\max(h) - \min(h)}{\mu - \min(h)}$.

While this linear approach offers some control, it has two key limitations: (1) the output range $R$ is upper-bounded by the non-negativity requirement, and (2) the linear mapping does not allow for "curvature" control to amplify or compress differences between entropy values.

\textbf{\textit{Attempt 3 (Final).}} To address the non-negativity constraint, we propose mapping the transformed values into the exponential space by applying a logarithmic transformation to the entropy values: 
\begin{equation}
    f(h)=e^{a\ln{h}+b}=h^a \cdot e^b,
\end{equation}

where exponent parameter $a$ controls the curvature of the transformation, providing flexibility in how the entropy values are reshaped. This formulation automatically ensures $f(h)$ is non-negative for $h>0$. Substituting into the normalization constraint and solving for $b$:
\begin{equation*}
\frac{1}{N} \sum_{i=1}^N f(h_i) = \frac{1}{N} \sum_{i=1}^N h_i^{\,a}\, e^{\,b} = e^{\,b} \,\frac{1}{N} \sum_{i=1}^N h_i^{\,a} = 1
\Rightarrow e^{\,b} = \frac{N}{\sum_{i=1}^N h_i^{\,a}}
\Rightarrow b = \ln\!\Bigl(\frac{N}{\sum_{i=1}^N h_i^{\,a}}\Bigr).
\end{equation*}

Therefore, our final mapping function is:
\begin{equation}
    f(h) = h^{\,a} \,\frac{N}{\sum_{i=1}^N h_i^{\,a}}.
\end{equation}
The exponent parameter $a$ provides curvature control over the transformation: $f(h)$ enhances differences between weights when $a > 1$; $f(h)$ compresses differences when $0 < a < 1$; $f(h)$ inverts the weight distribution when $a < 0$.
\input{table/algorithm}
\subsection{Loss Objective} 

The resulting weight is then integrated into the loss objective as:
\begin{equation}
\label{eq:w}
    \mathcal{L}_{\mathrm{EAST}}(\theta) = f(h) \cdot \mathcal{L}(\theta), \quad \text{where } h = H(x)
\end{equation}
where \( f(h) \) is the mapping function applied to the entropy value \( H(x) \), and \( \mathcal{L}(\theta) \) denotes the base loss. EAST is flexible and can be seamlessly applied to various loss functions(e.g., SFT or DPO). Furthermore, it naturally supports iterative training by repeating the weighting and fine-tuning process. The full procedure is detailed in Algorithm~\ref{alg:east}.

%% file: table/algorithm.tex

    

    
 


\begin{algorithm}[ht]
\caption{Entropy-Based Adaptive Weighting for Self-Training(EAST)}
\label{alg:east}
\KwIn{Initial model parameters $\theta$, training data $\mathcal{D} = \{(x_i, y_i)\}_{i=1}^{N}$, exponent parameter $a$, maximum iterations $T$}
\KwOut{Trained model parameters $\hat{\theta}$}

\For{$t = 1$ \KwTo $T$}{
    \ForEach{question $x_i \in \mathcal{D}$}{
        Generate $n$ responses and cluster them into $k_i$ groups by final answers with proportions $p_1, \dots, p_{k_i}$ per group\;
        Compute entropy: $h_i = -\sum_{j=1}^{k_i} p_j \log p_j$\;
    }
    Compute coefficient: $e^b = \frac{N}{\sum_{i=1}^{N} h_i^a}$\;
    
    \ForEach{question $x_i \in \mathcal{D}$}{
        Compute weight: $f(h_i) = h_i^a \cdot e^b$\;
        
        Update loss function: $\mathcal{L}_{\mathrm{EAST}}(\theta; x_i) = f(h_i) \cdot \mathcal{L}(\theta; x_i)$\;
    }
    Train model by minimizing $\mathcal{L}_{\mathrm{EAST}}$\;
}
\Return{$\hat{\theta}$}
\end{algorithm}

%% file: section/experiment.tex
\section{Experiment}
In this section, we present the experiment setup in Section~\ref{sec:setting} and main results in Section~\ref{sec:result}. Then, we provide further ablation study in Section~\ref{sec:ablation}. 
\subsection{Experiment Setup}
\label{sec:setting}

\textbf{Dataset.} We evaluate EAST on two mathematical benchmarks: MATH~\citep{hendrycks2021measuring} and GSM8K~\citep{cobbe2021training}. For training data, we prompt the backbone model to generate 128 samples per question and randomly select a positive–negative pair based on answer correctness, with the positive drawn from correct answers and the negative from incorrect ones. For evaluation, we note minor performance variations with vLLM across GPU types. For reproducibility and fair comparison, all results use the same GPU with temperature 0. We adapt the evaluation pipeline of \citet{yang2024qwen2}.

\textbf{Baseline}. We evaluate EAST across three loss functions: SFT, DPO\citep{rafailov2024direct}, and KTO\citep{ethayarajh2024kto}, which correspond to learning from positive samples, paired samples, and unpaired samples. In addition to the vanilla method, we incorporate weighting baselines that capture local uncertainty information. Specifically, local uncertainty information refers to model uncertainty derived exclusively from the token-level probabilities of a single selected sample response. This metric captures the uncertainty within an individual response, without accounting for the full distribution of all generated responses for a given question. One baseline uses the perplexity score for local information weighting (denoted as \textbf{LW(P)}), which is normalized within each batch to ensure stability and fair comparison. Another baseline (denoted as \textbf{LW(W)}) is inspired by WPO~\citep{zhou2024wpo}, which computes adaptive weights based on the log-likelihood of the sample response.
Detailed formulations for both baselines are provided in the Appendix~\ref{appendix:baseline}.


\textbf{Model Configuration.} We conduct experiments systematically based on two backbone models: \texttt{Llama-3.2-1B-Instruct} and \texttt{Llama-3.1-8B-Instruct}. For SFT, we use a learning rate of 2e-6 for 1B on GSM8K and MATH datasets. We adapt LoRA for 8B model with learning rate as 5e-5 for GSM8K and 2e-5 for MATH. For DPO, we adapt a learning rate of 2e-7 for 1B with $\beta=0.01$ and 2e-6 for 8B using LoRA with $\beta=0.1$ for both datasets. For KTO, we adapt a learning rate of 2e-7 for 1B with $\beta=0.05$ and 2e-6 for 8B using LoRA with $\beta=0.1$ for both datasets. For both baselines and EAST, we use the same set of hyperparameters as the vanilla method to ensure fair comparison. Each model is trained for three epochs with a batch size of 16 and warmup ratio of 0.1. We adapt the exponent parameters $a$ in the range $[-3,3]$ to fully investigate the functionality of the mapping function. Detailed hyperparameter study is provided in Appendix~\ref{appendix:setup}. 

\subsection{Experiment Results}
\label{sec:result}

\input{table/main_exp}

\begin{figure}[b]
\vspace{-0.6em}
    \centering
        \includegraphics[width=\linewidth]{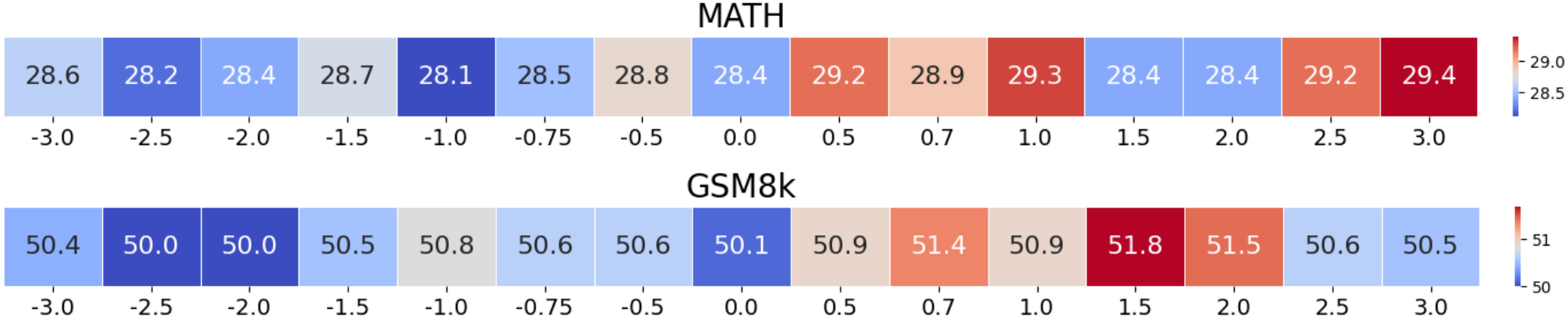}
        \vspace{-1.3em}
            \caption{Performance(accuracy (\%)) of various exponent parameters $a$ on GSM8K and MATH datasets using LLaMA-3.2-1B.}

    \label{fig:sft}
    \vspace{-0.2em}
\end{figure}

\begin{figure}[H]
    \centering
    \includegraphics[width=0.85\linewidth]{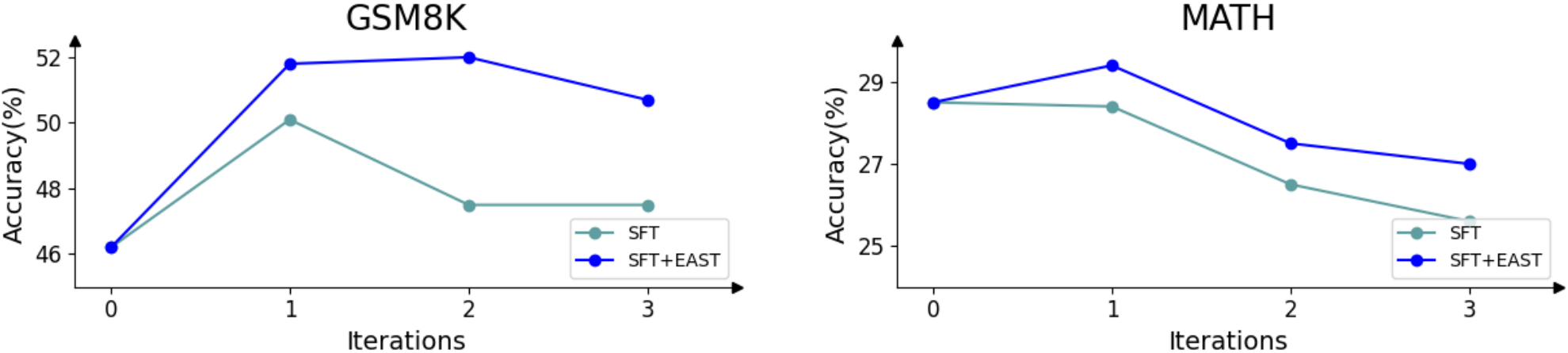}
    
    \vspace{-0.2em}
    \caption{Comparison of iterative learning performance (accuracy (\%)) between vanilla SFT and EAST on LLaMA-3.2-1B.  }
    \label{fig:iterative}
    \vspace{-0.2em}
\end{figure}

Table~\ref{table:main} presents the performance in terms of accuracy score of EAST compared to the vanilla method and baselines on the GSM8K and MATH benchmarks during the first iteration. The impact of different exponent parameters $a$ is shown in Figure~\ref{fig:sft}. Additionally, experiment results from iterative learning are presented in Figure~\ref{fig:iterative}. We have the following observations:
\newpage
\textbf{Observation 1: EAST outperforms the vanilla method and baselines.} As shown in Table~\ref{table:main}, EAST consistently improves performance compared to the vanilla method and baselines across both benchmarks under all loss functions. The vanilla SFT method struggles to outperform the default backbone model when trained on self-generated data, especially on the challenging MATH dataset. For instance, SFT achieves accuracies of 28.4\% and 50.0\% on MATH using LLaMA-3.2-1B and LLaMA-3.1-8B, respectively, which are slightly lower than the corresponding default model performances of 28.5\% and 50.4\%. In comparison, EAST improves the results to 29.4\% and 51.2\%, demonstrating over a 1\% absolute gain relative to SFT by focusing on more informative training examples. A similar trend is also observed on KTO: the vanilla KTO achieves only 83.9\% and 48.9\% on the 8B model, while EAST boosts the performance to 85.1\% and 51.0\%, representing gains of 1.2\% and 2.1\%, respectively.
Integrating EAST with the DPO loss function also leads to consistent gains. For example, on GSM8K with the LLaMA-3.2-1B model, EAST improves performance from 50.2\% to 51.9\%.

We also evaluate baselines (LW(W) and LW(P)) that use local uncertainty information of model. These approaches rely on the likelihood of next-token prediction for a given sample response, rather than the overall sample distribution, which also accounts for the correctness of the sample path. Results show that EAST outperforms both local information baselines across all loss functions and benchmarks. Notably, on the MATH dataset using the LLaMA-3.2-1B model, local weighting baselines achieve only 28.1\% (LW(W)) and 28.4\% (LW(P)), which is lower than both the vanilla DPO (28.7\%) and the default model (28.5\%). In contrast, EAST achieves a significantly higher score of 29.7\%, suggesting that local weighting may be more sensitive to token-level noise and  potentially limiting its training effectiveness.

\textbf{Observation 2: Weighting more on uncertain data contributes to performance improvement. } Figure~\ref{fig:sft} demonstrates the accuracy score of different parameters $a$ using SFT method on both benchmarks using LLaMA-3.2-1B model. The figure demonstrates that while performance varies with different values of \( a \), the best results are achieved when \( a > 0 \) for both datasets. The model achieves peak accuracy on the GSM8K dataset at $a = 1.5$ with 51.8\%, compared to 50.6\% and 50.8\% when $a = -0.75$ and $a = -1$, respectively. Similarly, for the MATH dataset, performance reaches 29.3\% at $a = 1$ and 29.4\% at $a = 3$, outperforming the 28.1\% and 28.5\% observed at $a = -1$ and $a = -0.75$, respectively. These results suggest that prioritizing uncertain data helps the model enhance its reasoning ability during training, leading to improved performance. 


\textbf{Observation 3: EAST demonstrates consistent benefits in iterative training.} To further investigate the performance of EAST in iterative learning, we conduct experiments with iterations $T=3$ using LLaMA-3.2-1B on both the MATH and GSM8K datasets, with results presented in Figure~\ref{fig:iterative} . EAST consistently outperforms vanilla SFT across iterations on both datasets. Notably, EAST maintains strong performance over time, while vanilla SFT appears to overfit on self-generated data in GSM8K. Although both methods struggle with iterative learning on the MATH dataset, EAST still demonstrates a relative advantage.


\subsection{Ablation Study: Effect of Accuracy-Based and Reject-Based Weights}
\label{sec:ablation}
\input{table/ablation}

To further investigate the effectiveness of entropy-based weighting, we compare it to alternative weighting strategies grounded in other distributional metrics. Noting that entropy is typically low when a single answer—whether correct or incorrect—dominates the distribution, we explore two complementary approaches: \textit{accuracy-based weighting}, which considers the proportion of the correct answer, and \textit{rejection-based weighting}, which measures the dominance of the most frequent incorrect answer.

\textbf{Accuracy-Based Weights.} Accuracy-based weighting leverages the accuracy ratio of model for each question to determine the corresponding weight. Specifically, in the self-training pipeline with $n$ samples for each question in the training data and the accuracy score is computed based on the proportion of correct predictions: $A(x) = \frac{1}{n}\sum_{i=1}^{n} \mathbb{1}(y_i = y^*)$, where $y_i$ represents the i-th sampled prediction and $y^*$ denotes the ground truth. For notational simplicity, let $s_i = 1 - A(x_i)$ represent the inverse accuracy score for question $i$ and the weight is aggregated using mapping function $f(s_i)$, as detailed in Section~\ref{sec:mapping}. Intuitively, when $s_i$ is large, the model faces challenges when solving the problems.

\textbf{Rejected-Based Weight.} Recent studies indicate that large language models (LLMs) struggle with self-correction, particularly when they generate responses with high confidence~\citep{kumar2024training, yang2024confidence}. To further investigate this phenomenon, we propose a novel weighting scheme that prioritizes the most “stubborn” questions—those for which the model repeatedly produces the same incorrect answers. Specifically, for all samples that yield incorrect answers, we partition them into $\hat{k}$ clusters, where each cluster corresponds to a distinct final answer. Next, we calculate the proportion $p$ of each incorrect answer cluster and identify the most frequent (dominant) mistake: $R(x) = \max_{j\in[\hat{k}]} p_j$.
For notational simplicity, let $r_i = R(x_i)$ represent the inverse accuracy score for question $i$ and the weight is aggregated using mapping function $f(r_i)$, as detailed in Section~\ref{sec:mapping}.
\begin{figure}[t]
\vspace{-1em}
    \centering
        \includegraphics[width=\linewidth]{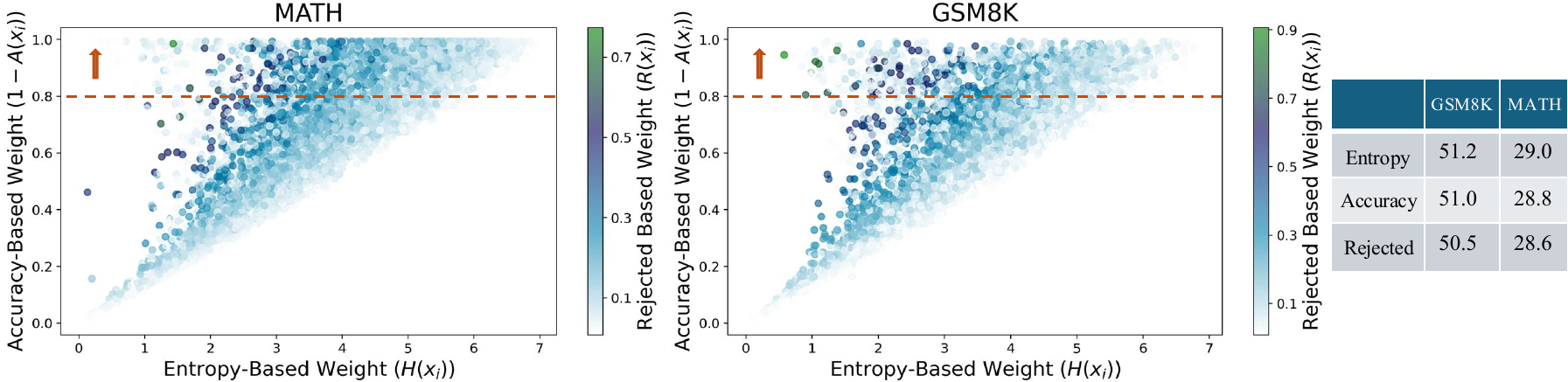}
    \vspace{-0.7em}
    \caption{The figure illustrates the distribution of training data in entropy-based, accuracy-based, and rejected-based values. Each point represents a training example ($x_i$), with coordinates ($H(x_i), 1-A(x_i)$) for entropy-based and accuracy-based values, and color indicating the rejected-based value ($R(x_i)$). The accompanying table reports the performance (accuracy(\%)) of three weighting strategies on the GSM8K and MATH datasets. }

    \label{fig:ablation}
    \vspace{-1em}
\end{figure}

\textbf{Experiment Result and Analysis.} As shown in Equation~\ref{eq:entropy}, both $A(x_i)$ and $R(x_i)$ can be interpreted as components of the probability distribution $H(x_i)$ over predicted answers. According to the equation, when either $A(x_i)$ or $R(x_i)$ is large, the entropy $H(x_i)$ tends to be low, reflecting greater certainty in the model’s predictions. This relationship is further illustrated in Figure~\ref{fig:ablation}, where higher entropy values are associated with larger $(1 - A(x_i))$ (i.e., lower accuracy) and smaller $R(x_i)$ values (i.e., greater diversity in incorrect predictions). However, when $(1 - A(x_i))$ becomes large—exceeding 0.8, for instance—this does not necessarily imply a low $R(x_i)$. In fact, Figure~\ref{fig:ablation} shows that many of these low-accuracy samples still exhibit high $R(x_i)$ values, indicating repeated, confident errors. As a result, applying the mapping function $f$ to such cases may overemphasize these “stubborn” questions during training, potentially skewing the learning dynamics and degrading overall performance. In contrast, entropy-based weighting effectively addresses this problem by automatically assigning lower weights to cases where a single incorrect answer dominates. 

Empirical results of the three weighting strategies on both datasets are reported in Figure~\ref{fig:ablation}, using SFT and LLaMA-3.2-1B. The results show that entropy-based weighting outperforms other strategies on both datasets. In contrast, reject-based weighting consistently yields the lowest performance across both benchmarks, while accuracy-based weighting achieves comparable results but exhibits certain limitations.



%% file: table/main_exp.tex
\begin{table*}[t]
\vspace{-1.8em}
\caption{Experimental results in terms of accuracy(\%) on GSM8K and MATH benchmarks. The best performance under each loss category is highlighted in \textbf{bold}. Significant boosts ($\ge$ 1\%) of EAST over both the vanilla method and baselines are \underline{underlined}. }
\label{table:main}
\centering
\SetTblrInner{rowsep=0.95pt}
\resizebox{0.98\textwidth}{!}{%
\begin{tblr}{
  colspec = {lcccccc},
  row{1-2} = {bg=gray!25},        
  row{4-7, 12-15} = {bg=gray!10},  
}
\toprule
\SetCell[r=2]{l}{Setting} & \SetCell[c=3]{c}{\texttt{LLaMA-3.2-1B}} & & & \SetCell[c=3]{c}{\texttt{LLaMA-3.1-8B}} & & \\
\cmidrule[lr]{2-4}\cmidrule[lr]{5-7}
& GSM8K$(\%)$ & MATH$(\%)$ & AVG$(\%)$ & GSM8K$(\%)$ & MATH$(\%)$ & AVG$(\%)$ \\
\midrule
\textit{default} & 46.2 & 28.5 & 37.3 & 82.8 & 50.4 & 66.6 \\
SFT              & 50.1 & 28.4 & 39.2 & 85.0 & 50.0 & 67.5 \\
+LW(W)       & 50.9 & 28.5 & 39.7 & 84.8 & 50.9 & 67.8 \\
+LW(P)       & 51.2& 28.4&  39.8&  85.1&  50.8& 68.0 \\
+EAST             & \textbf{51.8} & \textbf{29.4} & \textbf{40.6} & \underline{\textbf{86.1}} & \textbf{51.2} & \textbf{68.6} \\
DPO              & 50.2 & 28.7 & 39.5 & 84.6 & 50.1 & 67.5 \\
+LW(W)      & 50.9 & 28.1 & 39.5 & 85.1 & 50.2 & 67.6 \\
+LW(P)       & 50.3&28.4& 39.4 & 85.2 &  50.8& 68.0 \\
+EAST             & \underline{\textbf{51.9}} & \underline{\textbf{29.7}} & \underline{\textbf{40.8}} & \textbf{85.4} & \textbf{50.9} & \textbf{68.1} \\
KTO              & \textbf{53.0} & 28.8 & 40.9 & 83.9 & 48.9 & 66.4 \\
+LW(W)      & 52.9 & 28.2 & 40.6 &  84.3&  49.1&  66.7\\
+LW(P)       & 52.9& 28.9& 40.9 & 83.9 & 49.1 & 66.5 \\
+EAST             & \textbf{53.0} & \underline{\textbf{29.9}} & \textbf{41.5} & \textbf{85.1}&  \underline{\textbf{51.0}}& \underline{\textbf{68.1}} \\
\bottomrule
\end{tblr}%
}
\vspace{-0.5em}
\end{table*}

%% file: table/ablation.tex




%% file: section/related_work.tex
\section{Related Work}
\textbf{Self-Training on Mathematical Reasoning.}
Mathematical reasoning has emerged as a critical evaluation benchmark for Large Language Models (LLMs), as it directly correlates with logical reasoning capabilities and provides clear assessment metrics~\citep{azerbayev2023llemma,wang2023scibench, zhang2024mathverse, gao2024omni, liu2024mathbench}. Traditional methods rely on carefully curated manual datasets as demonstrations for fine-tuning~\citep{yue2023mammoth, yu2023metamath, luo2023wizardmath}.
As high-quality annotated data are expensive, numerous studies leverage rephrasing methods to augment datasets~\citep{deng2023rephrase,yu2023metamath}, or employ strong LLMs to generate synthetic data for knowledge distillation~\citep{taori2023stanford,chiang2023vicuna,ho2022large, fu2023specializing, gou2023tora}. 

Recently, several studies have explored using the target model to generate training data and enhance its performance through self-training~\citep{zelikman2022star, singh2023beyond,hosseini2024v, yu2023teaching, chen2024self, kumar2024training, tao2024survey}, while others extend such techniques for pair-wise alignment methods by leveraging negative samples generated from previous iterations~\citep{tajwar2024preference, xu2024dpo,sun2024easy,zhong2024dpo,ivison2024unpacking, xiong2024iterative, xie2024memorization, pang2025iterative}. For instance, \citet{sun2024easy} compare REST-EM with iterative DPO in a self-training pipeline, and \citet{xiong2024iterative} employs the multi-turn reasoning path for iterative learning process. For this work, we further optimize self-generated data usage by incorporating weighting strategies to improve reasoning capabilities.

\textbf{Alignment Method.} Reinforcement Learning from Human Feedback (RLHF) has emerged as an essential framework for aligning machine learning models with human preferences, emphasizing the importance of post-training optimization. Direct Preference Optimization (DPO) is a widely recognized method for alignment\citep{rafailov2024direct}. Recent studies have explored derivation of DPO \citep{xu2024contrastive,meng2024simpo,azar2024general}. For example, KTO directly maximizes the utility of generated outputs instead of focusing on the log-likelihood of preferences\citep{ethayarajh2024kto}. Some other studies focused on applying local weighting\citep{zhou2024wpo} or reward weighting upon DPO \citep{ adler2024nemotron, xiao2024comprehensive, yang2024weighted}. RPO incorporates reward gaps into preference learning to mitigate overfitting and better capture nuanced response quality\citep{adler2024nemotron}. However, it relies on an external reward model to assign weights to preference pairs. In contrast, WPO reweights preference pairs based on their likelihood under the current policy \citep{zhou2024wpo}. Nevertheless, WPO relies solely on local information of the given sample response without considering the overall sample distribution or controlling weight distribution skewness. 



%% file: section/conclusion.tex
\section{Conclusion}
This paper presents EAST, an entropy-based adaptive weighting method designed to emphasize uncertain data to improve reasoning capabilities during self-training. Through a tunable mapping function, EAST adjusts the degree of weighting applied to uncertain data. Experiments on the GSM8K and MATH benchmarks show consistent performance gains, demonstrating the effectiveness of the proposed method. These findings underscore the potential of adaptive weighting in enhancing reasoning capabilities and suggest directions for more effective self-training strategies in future research.

%% file: section/appendix.tex
\clearpage
\appendix
\begin{center}
	{\Large \textbf{Supplementary Material for EAST}}
\end{center}
\section{Reproducibility}
All code will be publicly available in the GitHub. All results are evaluated using the NVIDIA RTX A6000 GPU, following the evaluation pipeline of \citet{yang2024qwen2}.
\section{Experiment Setup}
\subsection{Baseline}
\label{appendix:baseline}
For local uncertainty information weighting, we use the standard perplexity(LW(P)):
\begin{equation}
\text{PPL}(x, y) = \exp\left( -\frac{1}{|y|} \sum_{t=1}^{|y|} \log \pi_\theta(y_t \mid x, y_{<t}) \right),
\end{equation}
where \( \pi_\theta(y_t \mid x, y_{<t}) \) denotes the model's predicted probability of token \( y_t \) conditioned on the input \( x \) and the preceding tokens \( y_{<t} \). We further normalize the perplexity score within each batch by dividing by the batch mean:
\begin{equation}
\widetilde{\text{PPL}}(x, y) = \frac{\text{PPL}(x, y)}{\frac{1}{B} \sum_{i=1}^{B} \text{PPL}(x^{(i)}, y^{(i)})},
\end{equation}
where \( B \) denotes the batch size, and \( \text{PPL}(x^{(i)}, y^{(i)}) \) is the perplexity of the \( i \)-th sample in the batch.

For local uncertainty information weighting, we use the formulation from WPO~\citep{zhou2024wpo} as our local weighting strategy:
\begin{equation}
w(x, y) = \exp\left( \frac{1}{|y|} \sum_{t=1}^{|y|} \log \frac{\pi_\theta(y_t \mid x, y_{<t})}{\sum_{v \in \mathcal{V}} \pi_\theta(v \mid x, y_{<t})^2} \right),
\end{equation}
For DPO, local weights are computed for both positive and negative pairs and multiplied to obtain the final weight, whereas for SFT, only the positive samples are used.

\subsection{Hyperparameter Study}
\label{appendix:setup}
For SFT, the learning rate in \texttt{Llama-3.2-1B-Instruct} is chosen from \(\{2\mathrm{e}{-6}, 5\mathrm{e}{-6}, 7\mathrm{e}{-6}, 1\mathrm{e}{-5}\}\), and in \texttt{Llama-3.1-8B-Instruct} from \(\{2\mathrm{e}{-5}, 5\mathrm{e}{-5}, 7\mathrm{e}{-5}, 1\mathrm{e}{-4}\}\).

For DPO and KTO, we tune the temperature parameter \(\beta\) within the set \(\{0.01, 0.05, 0.1\}\). In \texttt{Llama-3.2-1B-Instruct}, we search the learning rate in \(\{2\mathrm{e}{-7}, 5\mathrm{e}{-7}, 7\mathrm{e}{-7}, 1\mathrm{e}{-6}\}\), while for \texttt{Llama-3.1-8B-Instruct}, the learning rate is selected from \(\{2\mathrm{e}{-6}, 5\mathrm{e}{-6}, 7\mathrm{e}{-6}, 1\mathrm{e}{-5}\}\). 

The baseline method and EAST share the same set of hyperparameters as the vanilla method to ensure a fair comparison. For EAST, we additionally search the exponent parameter \(a\) from the range \(\{-3, -2.5, -2, -1.5, -1.25, -1, -0.5, 0.1, 0.2, 0.5, 0.7, 1, 1.5, 2, 2.5, 3\}\).

For \texttt{Llama-3.1-8B-Instruct}, we apply LoRA with a rank of 16 and a LoRA alpha of 16. All models are trained using \texttt{bf16} precision, and we use the \texttt{AdamW} optimizer.

For the ablation study, we report the average accuracy scores for $a \in \{0.5, 1, 1.5\}$ across all three weighting methods.

%% file: colm2025_conference.bbl
\begin{thebibliography}{46}
\providecommand{\natexlab}[1]{#1}
\providecommand{\url}[1]{\texttt{#1}}
\expandafter\ifx\csname urlstyle\endcsname\relax
  \providecommand{\doi}[1]{doi: #1}\else
  \providecommand{\doi}{doi: \begingroup \urlstyle{rm}\Url}\fi

\bibitem[Adler et~al.(2024)Adler, Agarwal, Aithal, Anh, Bhattacharya, Brundyn, Casper, Catanzaro, Clay, Cohen, et~al.]{adler2024nemotron}
Bo~Adler, Niket Agarwal, Ashwath Aithal, Dong~H Anh, Pallab Bhattacharya, Annika Brundyn, Jared Casper, Bryan Catanzaro, Sharon Clay, Jonathan Cohen, et~al.
\newblock Nemotron-4 340b technical report.
\newblock \emph{arXiv preprint arXiv:2406.11704}, 2024.

\bibitem[Azar et~al.(2024)Azar, Guo, Piot, Munos, Rowland, Valko, and Calandriello]{azar2024general}
Mohammad~Gheshlaghi Azar, Zhaohan~Daniel Guo, Bilal Piot, Remi Munos, Mark Rowland, Michal Valko, and Daniele Calandriello.
\newblock A general theoretical paradigm to understand learning from human preferences.
\newblock In \emph{International Conference on Artificial Intelligence and Statistics}, pp.\  4447--4455. PMLR, 2024.

\bibitem[Azerbayev et~al.(2023)Azerbayev, Schoelkopf, Paster, Santos, McAleer, Jiang, Deng, Biderman, and Welleck]{azerbayev2023llemma}
Zhangir Azerbayev, Hailey Schoelkopf, Keiran Paster, Marco~Dos Santos, Stephen McAleer, Albert~Q Jiang, Jia Deng, Stella Biderman, and Sean Welleck.
\newblock Llemma: An open language model for mathematics.
\newblock \emph{arXiv preprint arXiv:2310.10631}, 2023.

\bibitem[Chen et~al.(2024)Chen, Deng, Yuan, Ji, and Gu]{chen2024self}
Zixiang Chen, Yihe Deng, Huizhuo Yuan, Kaixuan Ji, and Quanquan Gu.
\newblock Self-play fine-tuning converts weak language models to strong language models.
\newblock \emph{arXiv preprint arXiv:2401.01335}, 2024.

\bibitem[Chiang et~al.(2023)Chiang, Li, Lin, Sheng, Wu, Zhang, Zheng, Zhuang, Zhuang, Gonzalez, et~al.]{chiang2023vicuna}
Wei-Lin Chiang, Zhuohan Li, Zi~Lin, Ying Sheng, Zhanghao Wu, Hao Zhang, Lianmin Zheng, Siyuan Zhuang, Yonghao Zhuang, Joseph~E Gonzalez, et~al.
\newblock Vicuna: An open-source chatbot impressing gpt-4 with 90\%* chatgpt quality.
\newblock \emph{See https://vicuna. lmsys. org (accessed 14 April 2023)}, 2\penalty0 (3):\penalty0 6, 2023.

\bibitem[Cobbe et~al.(2021)Cobbe, Kosaraju, Bavarian, Chen, Jun, Kaiser, Plappert, Tworek, Hilton, Nakano, et~al.]{cobbe2021training}
Karl Cobbe, Vineet Kosaraju, Mohammad Bavarian, Mark Chen, Heewoo Jun, Lukasz Kaiser, Matthias Plappert, Jerry Tworek, Jacob Hilton, Reiichiro Nakano, et~al.
\newblock Training verifiers to solve math word problems.
\newblock \emph{arXiv preprint arXiv:2110.14168}, 2021.

\bibitem[Deng et~al.(2023)Deng, Zhang, Chen, and Gu]{deng2023rephrase}
Yihe Deng, Weitong Zhang, Zixiang Chen, and Quanquan Gu.
\newblock Rephrase and respond: Let large language models ask better questions for themselves.
\newblock \emph{arXiv preprint arXiv:2311.04205}, 2023.

\bibitem[Ethayarajh et~al.(2024)Ethayarajh, Xu, Muennighoff, Jurafsky, and Kiela]{ethayarajh2024kto}
Kawin Ethayarajh, Winnie Xu, Niklas Muennighoff, Dan Jurafsky, and Douwe Kiela.
\newblock Kto: Model alignment as prospect theoretic optimization.
\newblock \emph{arXiv preprint arXiv:2402.01306}, 2024.

\bibitem[Fu et~al.(2023)Fu, Peng, Ou, Sabharwal, and Khot]{fu2023specializing}
Yao Fu, Hao Peng, Litu Ou, Ashish Sabharwal, and Tushar Khot.
\newblock Specializing smaller language models towards multi-step reasoning.
\newblock In \emph{International Conference on Machine Learning}, pp.\  10421--10430. PMLR, 2023.

\bibitem[Gao et~al.(2024)Gao, Song, Yang, Cai, Miao, Dong, Li, Ma, Chen, Xu, et~al.]{gao2024omni}
Bofei Gao, Feifan Song, Zhe Yang, Zefan Cai, Yibo Miao, Qingxiu Dong, Lei Li, Chenghao Ma, Liang Chen, Runxin Xu, et~al.
\newblock Omni-math: A universal olympiad level mathematic benchmark for large language models.
\newblock \emph{arXiv preprint arXiv:2410.07985}, 2024.

\bibitem[Gou et~al.(2023)Gou, Shao, Gong, Yang, Huang, Duan, Chen, et~al.]{gou2023tora}
Zhibin Gou, Zhihong Shao, Yeyun Gong, Yujiu Yang, Minlie Huang, Nan Duan, Weizhu Chen, et~al.
\newblock Tora: A tool-integrated reasoning agent for mathematical problem solving.
\newblock \emph{arXiv preprint arXiv:2309.17452}, 2023.

\bibitem[Hendrycks et~al.(2021)Hendrycks, Burns, Kadavath, Arora, Basart, Tang, Song, and Steinhardt]{hendrycks2021measuring}
Dan Hendrycks, Collin Burns, Saurav Kadavath, Akul Arora, Steven Basart, Eric Tang, Dawn Song, and Jacob Steinhardt.
\newblock Measuring mathematical problem solving with the math dataset.
\newblock \emph{arXiv preprint arXiv:2103.03874}, 2021.

\bibitem[Ho et~al.(2022)Ho, Schmid, and Yun]{ho2022large}
Namgyu Ho, Laura Schmid, and Se-Young Yun.
\newblock Large language models are reasoning teachers.
\newblock \emph{arXiv preprint arXiv:2212.10071}, 2022.

\bibitem[Hosseini et~al.(2024)Hosseini, Yuan, Malkin, Courville, Sordoni, and Agarwal]{hosseini2024v}
Arian Hosseini, Xingdi Yuan, Nikolay Malkin, Aaron Courville, Alessandro Sordoni, and Rishabh Agarwal.
\newblock V-star: Training verifiers for self-taught reasoners.
\newblock \emph{arXiv preprint arXiv:2402.06457}, 2024.

\bibitem[Huang et~al.(2022)Huang, Gu, Hou, Wu, Wang, Yu, and Han]{huang2022large}
Jiaxin Huang, Shixiang~Shane Gu, Le~Hou, Yuexin Wu, Xuezhi Wang, Hongkun Yu, and Jiawei Han.
\newblock Large language models can self-improve.
\newblock \emph{arXiv preprint arXiv:2210.11610}, 2022.

\bibitem[Ivison et~al.(2024)Ivison, Wang, Liu, Wu, Pyatkin, Lambert, Smith, Choi, and Hajishirzi]{ivison2024unpacking}
Hamish Ivison, Yizhong Wang, Jiacheng Liu, Zeqiu Wu, Valentina Pyatkin, Nathan Lambert, Noah~A Smith, Yejin Choi, and Hannaneh Hajishirzi.
\newblock Unpacking dpo and ppo: Disentangling best practices for learning from preference feedback.
\newblock \emph{arXiv preprint arXiv:2406.09279}, 2024.

\bibitem[Kumar et~al.(2024)Kumar, Zhuang, Agarwal, Su, Co-Reyes, Singh, Baumli, Iqbal, Bishop, Roelofs, et~al.]{kumar2024training}
Aviral Kumar, Vincent Zhuang, Rishabh Agarwal, Yi~Su, John~D Co-Reyes, Avi Singh, Kate Baumli, Shariq Iqbal, Colton Bishop, Rebecca Roelofs, et~al.
\newblock Training language models to self-correct via reinforcement learning.
\newblock \emph{arXiv preprint arXiv:2409.12917}, 2024.

\bibitem[Li et~al.(2024)Li, Chen, Chen, Zhang, Su, Xing, and Zhang]{li2024confidence}
Loka Li, Zhenhao Chen, Guangyi Chen, Yixuan Zhang, Yusheng Su, Eric Xing, and Kun Zhang.
\newblock Confidence matters: Revisiting intrinsic self-correction capabilities of large language models.
\newblock \emph{arXiv preprint arXiv:2402.12563}, 2024.

\bibitem[Liu et~al.(2024)Liu, Zheng, Qiao, Duan, Fei, Zhou, Zhang, Zhang, Lin, and Chen]{liu2024mathbench}
Hongwei Liu, Zilong Zheng, Yuxuan Qiao, Haodong Duan, Zhiwei Fei, Fengzhe Zhou, Wenwei Zhang, Songyang Zhang, Dahua Lin, and Kai Chen.
\newblock Mathbench: Evaluating the theory and application proficiency of llms with a hierarchical mathematics benchmark.
\newblock \emph{arXiv preprint arXiv:2405.12209}, 2024.

\bibitem[Luo et~al.(2023)Luo, Sun, Xu, Zhao, Lou, Tao, Geng, Lin, Chen, and Zhang]{luo2023wizardmath}
Haipeng Luo, Qingfeng Sun, Can Xu, Pu~Zhao, Jianguang Lou, Chongyang Tao, Xiubo Geng, Qingwei Lin, Shifeng Chen, and Dongmei Zhang.
\newblock Wizardmath: Empowering mathematical reasoning for large language models via reinforced evol-instruct.
\newblock \emph{arXiv preprint arXiv:2308.09583}, 2023.

\bibitem[Luong et~al.(2024)Luong, Zhang, Jie, Sun, Jin, and Li]{luong2024reft}
Trung~Quoc Luong, Xinbo Zhang, Zhanming Jie, Peng Sun, Xiaoran Jin, and Hang Li.
\newblock Reft: Reasoning with reinforced fine-tuning.
\newblock \emph{arXiv preprint arXiv:2401.08967}, 2024.

\bibitem[Meng et~al.(2024)Meng, Xia, and Chen]{meng2024simpo}
Yu~Meng, Mengzhou Xia, and Danqi Chen.
\newblock Simpo: Simple preference optimization with a reference-free reward.
\newblock \emph{arXiv preprint arXiv:2405.14734}, 2024.

\bibitem[Pang et~al.(2025)Pang, Yuan, He, Cho, Sukhbaatar, and Weston]{pang2025iterative}
Richard~Yuanzhe Pang, Weizhe Yuan, He~He, Kyunghyun Cho, Sainbayar Sukhbaatar, and Jason Weston.
\newblock Iterative reasoning preference optimization.
\newblock \emph{Advances in Neural Information Processing Systems}, 37:\penalty0 116617--116637, 2025.

\bibitem[Rafailov et~al.(2024)Rafailov, Sharma, Mitchell, Manning, Ermon, and Finn]{rafailov2024direct}
Rafael Rafailov, Archit Sharma, Eric Mitchell, Christopher~D Manning, Stefano Ermon, and Chelsea Finn.
\newblock Direct preference optimization: Your language model is secretly a reward model.
\newblock \emph{Advances in Neural Information Processing Systems}, 36, 2024.

\bibitem[Saeidi et~al.(2024)Saeidi, Verma, and Baral]{saeidi2024insights}
Amir Saeidi, Shivanshu Verma, and Chitta Baral.
\newblock Insights into alignment: Evaluating dpo and its variants across multiple tasks.
\newblock \emph{arXiv preprint arXiv:2404.14723}, 2024.

\bibitem[Singh et~al.(2023)Singh, Co-Reyes, Agarwal, Anand, Patil, Liu, Harrison, Lee, Xu, Parisi, et~al.]{singh2023beyond}
Avi Singh, John~D Co-Reyes, Rishabh Agarwal, Ankesh Anand, Piyush Patil, Peter~J Liu, James Harrison, Jaehoon Lee, Kelvin Xu, Aaron Parisi, et~al.
\newblock Beyond human data: Scaling self-training for problem-solving with language models.
\newblock \emph{arXiv preprint arXiv:2312.06585}, 2023.

\bibitem[Sun et~al.(2024)Sun, Yu, Shen, Liu, Yang, Welleck, and Gan]{sun2024easy}
Zhiqing Sun, Longhui Yu, Yikang Shen, Weiyang Liu, Yiming Yang, Sean Welleck, and Chuang Gan.
\newblock Easy-to-hard generalization: Scalable alignment beyond human supervision.
\newblock \emph{arXiv preprint arXiv:2403.09472}, 2024.

\bibitem[Tajwar et~al.(2024)Tajwar, Singh, Sharma, Rafailov, Schneider, Xie, Ermon, Finn, and Kumar]{tajwar2024preference}
Fahim Tajwar, Anikait Singh, Archit Sharma, Rafael Rafailov, Jeff Schneider, Tengyang Xie, Stefano Ermon, Chelsea Finn, and Aviral Kumar.
\newblock Preference fine-tuning of llms should leverage suboptimal, on-policy data.
\newblock \emph{arXiv preprint arXiv:2404.14367}, 2024.

\bibitem[Tao et~al.(2024)Tao, Lin, Chen, Li, Wu, Li, Jin, Huang, Tao, and Zhou]{tao2024survey}
Zhengwei Tao, Ting-En Lin, Xiancai Chen, Hangyu Li, Yuchuan Wu, Yongbin Li, Zhi Jin, Fei Huang, Dacheng Tao, and Jingren Zhou.
\newblock A survey on self-evolution of large language models.
\newblock \emph{arXiv preprint arXiv:2404.14387}, 2024.

\bibitem[Taori et~al.(2023)Taori, Gulrajani, Zhang, Dubois, Li, Guestrin, Liang, and Hashimoto]{taori2023stanford}
Rohan Taori, Ishaan Gulrajani, Tianyi Zhang, Yann Dubois, Xuechen Li, Carlos Guestrin, Percy Liang, and Tatsunori~B Hashimoto.
\newblock Stanford alpaca: An instruction-following llama model, 2023.

\bibitem[Wang et~al.(2023)Wang, Hu, Lu, Zhu, Zhang, Subramaniam, Loomba, Zhang, Sun, and Wang]{wang2023scibench}
Xiaoxuan Wang, Ziniu Hu, Pan Lu, Yanqiao Zhu, Jieyu Zhang, Satyen Subramaniam, Arjun~R Loomba, Shichang Zhang, Yizhou Sun, and Wei Wang.
\newblock Scibench: Evaluating college-level scientific problem-solving abilities of large language models.
\newblock \emph{arXiv preprint arXiv:2307.10635}, 2023.

\bibitem[Xiao et~al.(2024)Xiao, Wang, Gan, Zhao, He, Tuan, Chen, Jiang, Zhao, and Wu]{xiao2024comprehensive}
Wenyi Xiao, Zechuan Wang, Leilei Gan, Shuai Zhao, Wanggui He, Luu~Anh Tuan, Long Chen, Hao Jiang, Zhou Zhao, and Fei Wu.
\newblock A comprehensive survey of datasets, theories, variants, and applications in direct preference optimization.
\newblock \emph{arXiv preprint arXiv:2410.15595}, 2024.

\bibitem[Xie et~al.(2024)Xie, Huang, Zhang, Yu, Chen, Lin, Li, Ghazi, and Kumar]{xie2024memorization}
Chulin Xie, Yangsibo Huang, Chiyuan Zhang, Da~Yu, Xinyun Chen, Bill~Yuchen Lin, Bo~Li, Badih Ghazi, and Ravi Kumar.
\newblock On memorization of large language models in logical reasoning.
\newblock \emph{arXiv preprint arXiv:2410.23123}, 2024.

\bibitem[Xiong et~al.(2024)Xiong, Dong, Ye, Wang, Zhong, Ji, Jiang, and Zhang]{xiong2024iterative}
Wei Xiong, Hanze Dong, Chenlu Ye, Ziqi Wang, Han Zhong, Heng Ji, Nan Jiang, and Tong Zhang.
\newblock Iterative preference learning from human feedback: Bridging theory and practice for rlhf under kl-constraint.
\newblock In \emph{Forty-first International Conference on Machine Learning}, 2024.

\bibitem[Xu et~al.(2024{\natexlab{a}})Xu, Sharaf, Chen, Tan, Shen, Van~Durme, Murray, and Kim]{xu2024contrastive}
Haoran Xu, Amr Sharaf, Yunmo Chen, Weiting Tan, Lingfeng Shen, Benjamin Van~Durme, Kenton Murray, and Young~Jin Kim.
\newblock Contrastive preference optimization: Pushing the boundaries of llm performance in machine translation.
\newblock \emph{arXiv preprint arXiv:2401.08417}, 2024{\natexlab{a}}.

\bibitem[Xu et~al.(2024{\natexlab{b}})Xu, Fu, Gao, Ye, Liu, Mei, Wang, Yu, and Wu]{xu2024dpo}
Shusheng Xu, Wei Fu, Jiaxuan Gao, Wenjie Ye, Weilin Liu, Zhiyu Mei, Guangju Wang, Chao Yu, and Yi~Wu.
\newblock Is dpo superior to ppo for llm alignment? a comprehensive study.
\newblock \emph{arXiv preprint arXiv:2404.10719}, 2024{\natexlab{b}}.

\bibitem[Yang et~al.(2024{\natexlab{a}})Yang, Yang, Hui, Zheng, Yu, Zhou, Li, Li, Liu, Huang, et~al.]{yang2024qwen2}
An~Yang, Baosong Yang, Binyuan Hui, Bo~Zheng, Bowen Yu, Chang Zhou, Chengpeng Li, Chengyuan Li, Dayiheng Liu, Fei Huang, et~al.
\newblock Qwen2 technical report.
\newblock \emph{arXiv preprint arXiv:2407.10671}, 2024{\natexlab{a}}.

\bibitem[Yang et~al.(2024{\natexlab{b}})Yang, Zhang, Wang, Xu, Lin, and Sui]{yang2024confidence}
Zhe Yang, Yichang Zhang, Yudong Wang, Ziyao Xu, Junyang Lin, and Zhifang Sui.
\newblock Confidence vs critique: A decomposition of self-correction capability for llms.
\newblock \emph{arXiv preprint arXiv:2412.19513}, 2024{\natexlab{b}}.

\bibitem[Yang et~al.(2024{\natexlab{c}})Yang, Wan, Zhong, Shi, and Quan]{yang2024weighted}
Ziyi Yang, Fanqi Wan, Longguang Zhong, Tianyuan Shi, and Xiaojun Quan.
\newblock Weighted-reward preference optimization for implicit model fusion.
\newblock \emph{arXiv preprint arXiv:2412.03187}, 2024{\natexlab{c}}.

\bibitem[Yu et~al.(2023{\natexlab{a}})Yu, Jiang, Shi, Yu, Liu, Zhang, Kwok, Li, Weller, and Liu]{yu2023metamath}
Longhui Yu, Weisen Jiang, Han Shi, Jincheng Yu, Zhengying Liu, Yu~Zhang, James~T Kwok, Zhenguo Li, Adrian Weller, and Weiyang Liu.
\newblock Metamath: Bootstrap your own mathematical questions for large language models.
\newblock \emph{arXiv preprint arXiv:2309.12284}, 2023{\natexlab{a}}.

\bibitem[Yu et~al.(2023{\natexlab{b}})Yu, Peng, Galley, Gao, and Yu]{yu2023teaching}
Xiao Yu, Baolin Peng, Michel Galley, Jianfeng Gao, and Zhou Yu.
\newblock Teaching language models to self-improve through interactive demonstrations.
\newblock \emph{arXiv preprint arXiv:2310.13522}, 2023{\natexlab{b}}.

\bibitem[Yue et~al.(2023)Yue, Qu, Zhang, Fu, Huang, Sun, Su, and Chen]{yue2023mammoth}
Xiang Yue, Xingwei Qu, Ge~Zhang, Yao Fu, Wenhao Huang, Huan Sun, Yu~Su, and Wenhu Chen.
\newblock Mammoth: Building math generalist models through hybrid instruction tuning.
\newblock \emph{arXiv preprint arXiv:2309.05653}, 2023.

\bibitem[Zelikman et~al.(2022)Zelikman, Wu, Mu, and Goodman]{zelikman2022star}
Eric Zelikman, Yuhuai Wu, Jesse Mu, and Noah Goodman.
\newblock Star: Bootstrapping reasoning with reasoning.
\newblock \emph{Advances in Neural Information Processing Systems}, 35:\penalty0 15476--15488, 2022.

\bibitem[Zhang et~al.(2024)Zhang, Jiang, Zhang, Lin, Guo, Qiu, Zhou, Lu, Chang, Gao, et~al.]{zhang2024mathverse}
Renrui Zhang, Dongzhi Jiang, Yichi Zhang, Haokun Lin, Ziyu Guo, Pengshuo Qiu, Aojun Zhou, Pan Lu, Kai-Wei Chang, Peng Gao, et~al.
\newblock Mathverse: Does your multi-modal llm truly see the diagrams in visual math problems?
\newblock \emph{arXiv preprint arXiv:2403.14624}, 2024.

\bibitem[Zhong et~al.(2024)Zhong, Feng, Xiong, Zhao, He, Bian, and Wang]{zhong2024dpo}
Han Zhong, Guhao Feng, Wei Xiong, Li~Zhao, Di~He, Jiang Bian, and Liwei Wang.
\newblock Dpo meets ppo: Reinforced token optimization for rlhf.
\newblock \emph{arXiv preprint arXiv:2404.18922}, 2024.

\bibitem[Zhou et~al.(2024)Zhou, Agrawal, Zhang, Indurthi, Zhao, Song, Xu, and Zhu]{zhou2024wpo}
Wenxuan Zhou, Ravi Agrawal, Shujian Zhang, Sathish~Reddy Indurthi, Sanqiang Zhao, Kaiqiang Song, Silei Xu, and Chenguang Zhu.
\newblock Wpo: Enhancing rlhf with weighted preference optimization.
\newblock \emph{arXiv preprint arXiv:2406.11827}, 2024.

\end{thebibliography}
